\documentclass[10pt,twocolumn,letterpaper]{article}

\usepackage[pagenumbers]{cvpr} 

\usepackage{graphicx}
\usepackage{booktabs}
\usepackage{multirow}
\usepackage{microtype}
\usepackage{colortbl}
\usepackage{makecell}
\usepackage{booktabs}
\usepackage[dvipsnames]{xcolor}
\usepackage[accsupp]{axessibility}  

\definecolor{wall_color}{RGB}{174, 199, 232}
\definecolor{floor_color}{RGB}{152, 223, 138}
\definecolor{door_color}{RGB}{214, 39, 40}
\definecolor{cabinet_color}{RGB}{31, 119, 180}

\definecolor{cvprblue}{rgb}{0.21,0.49,0.74}
\usepackage[pagebackref,breaklinks,colorlinks,allcolors=cvprblue]{hyperref}

\def\paperID{1302} 
\def\confName{CVPR}
\def\confYear{2025}

\definecolor{tabhighlight}{HTML}{e5e5e5}
\definecolor{lightgray}{gray}{0.6}

\title{Cross-Modal and Uncertainty-Aware Agglomeration for Open-Vocabulary 3D Scene Understanding}

\author{Jinlong Li$^{1,\dagger}$ ~~~~~~~ {Cristiano Saltori}$^{1}$ ~~~~~~~ {Fabio Poiesi}$^{2}$  ~~~~~~~  {Nicu Sebe$^{1}$} \\
$^1$ University of Trento \quad  \quad $^2$ Fondazione Bruno Kessler\\
}

\begin{document}
\maketitle

\newcommand\blfootnote[1]{%
\begingroup 
\renewcommand\thefootnote{}\footnote{#1}%
\addtocounter{footnote}{-1}%
\endgroup 
}
    {
        \blfootnote{
          $^\dagger$Corresponding author: tyronejinlongli@gmail.com.
    
    }
}

\begin{abstract}
   
   \noindent The lack of a large-scale 3D-text corpus has led recent works to distill open-vocabulary knowledge from vision-language models (VLMs). However, these methods typically rely on a single VLM to align the feature spaces of 3D models within a common language space, which limits the potential of 3D models to leverage the diverse spatial and semantic capabilities encapsulated in various foundation models. In this paper, we propose Cross-modal and Uncertainty-aware Agglomeration for Open-vocabulary 3D Scene Understanding dubbed \textbf{CUA-O3D}, the first model to integrate multiple foundation models—such as CLIP, DINOv2, and Stable Diffusion—into 3D scene understanding. We further introduce a deterministic uncertainty estimation to adaptively distill and harmonize the heterogeneous 2D feature embeddings from these models.
   Our method addresses two key challenges: (1) incorporating semantic priors from VLMs alongside the geometric knowledge of spatially-aware vision foundation models, and (2) using a novel deterministic uncertainty estimation to capture model-specific uncertainties across diverse semantic and geometric sensitivities, helping to reconcile heterogeneous representations during training. 
   Extensive experiments on ScanNetV2 and Matterport3D demonstrate that our method not only advances open-vocabulary segmentation but also achieves robust cross-domain alignment and competitive spatial perception capabilities. Project webpage: \href{https://tyroneli.github.io/CUA_O3D}{CUA-O3D}.
    
\end{abstract}

\section{Introduction}
\label{sec:intro}

3D scene understanding serves as a crucial perception component for a wide array of real-world applications to help models better understand the physical world, including robot navigation, autonomous vehicles, and virtual reality~\cite{graham20183d, vu2022softgroup, li2020lidar, behley2019iccv}. 
Typical approaches necessitate a dataset of semantically annotated point clouds, which is both time-consuming (\textit{e.g.}, 22.3 minutes for annotating a single scene with 20 classes~\cite{dai2017scannet}) and unable to encompass all possible existing categories, thereby limiting their practical utility. 
To address this limitation, recent works have explored the open-vocabulary 3D scene understanding setting~\cite{conceptfusion,peng2023openscene,ding2022language,Mei2024,Mei2025}, aiming to localize and recognize arbitrary object classes. 
This objective is achieved by leveraging Vision-Language Models (VLMs) \cite{radford2021learning,jia2021scaling} that, having been pre-trained on billions of image-text pairs~\cite{schuhmann2022laion}, can gauge the alignment between textual and visual inputs, facilitating a range of 2D open-vocabulary tasks~\cite{gu2021open,zhong2022regionclip,li2022languagedriven,ghiasi2022scaling,ko2023open}. 
To adapt these models for 3D downstream tasks (\textit{e.g.,}~semantic segmentation), various strategies, predominantly based on distilling multi-view 2D visual features into a 3D-specific model~\cite{ha2022semabs,conceptfusion,peng2023openscene,ding2022language,xu20243d}, have been employed.

\begin{figure}[t]
    \centering
    \includegraphics[width=\linewidth]{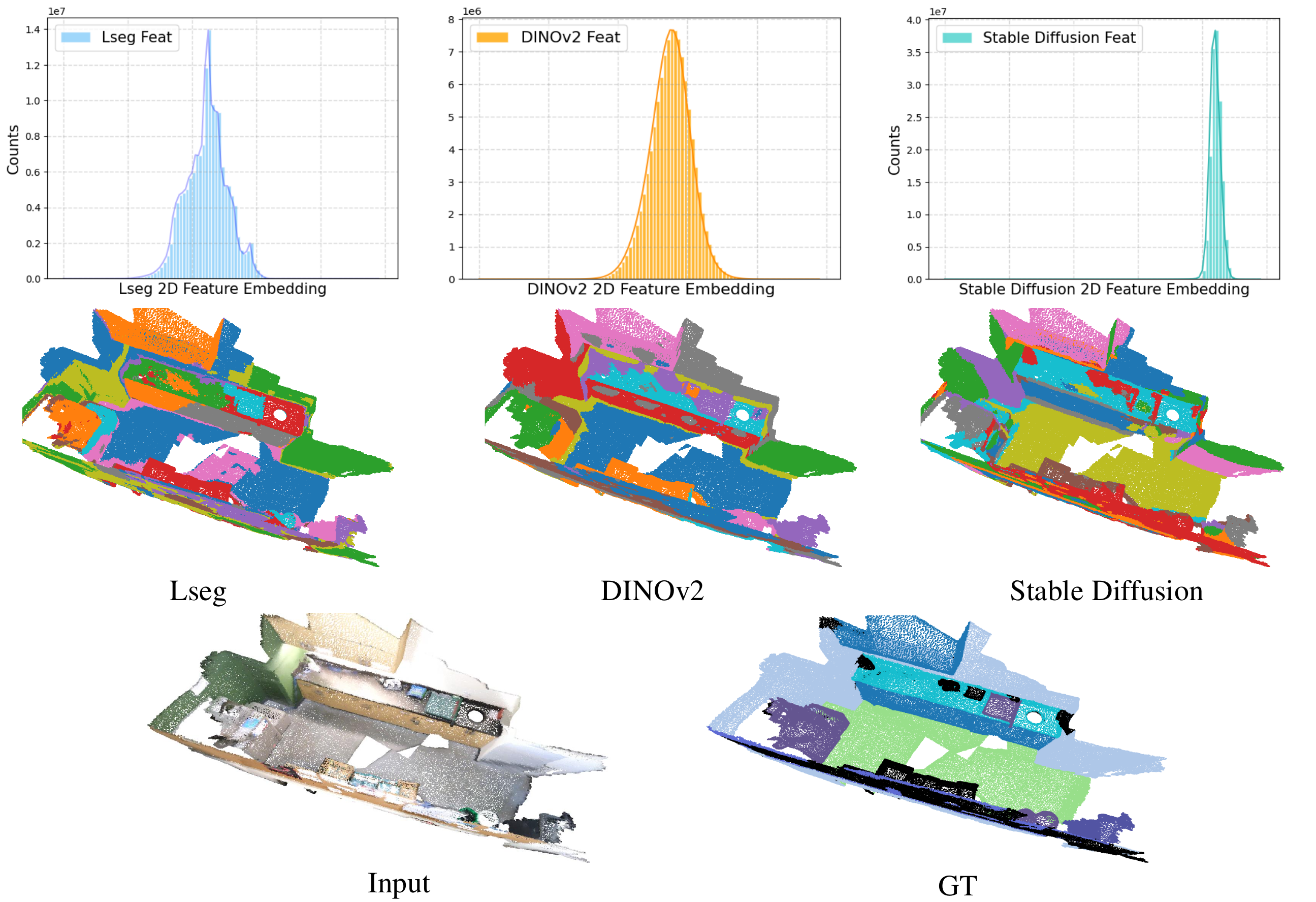}
    \vspace{-0.6cm}
    \caption{Top is feature distribution analysis of different 2d projected feature embeddings from various foundation models (Lseg, DINOv2 and Stable Diffusion), enumerating on the overall ScanNetV2 \textit{train} set and counting the frequency of all point features within each bin interval. Bottom is the sample utilizing \textit{K-Means} to cluster projected 3D features into specified clusters to make segmentation comparisons. Different foundation models illustrate heterogeneous yet complementary results.}
    \label{fig:motivation_fig}
    \vspace{-0.6cm}
\end{figure}

However, these methods mainly focus on distilling knowledge from one single VLM which limits the potential of 3D models to leverage the diverse spatial and semantic capabilities that have been trained on large-scale images or image-text pairs corpus, as shown in Table~\ref{tab:vfm}. Given the existence of various foundation models, such as CLIP~\cite{radford2021learning}, DINOv2~\cite{oquab2023dinov2} and Stable Diffusion~\cite{rombach2022high}, \textit{etc.}, there has been under-explored on how to make better utility of these 2D foundation models to develop the 3D foundation model, wherein no large-scale 3D point-cloud or 3D-text pair corpus available. Very recently, some works~\cite{el2024probing,man2024lexicon3d} have started probing the potential of these 2D foundation models on 3D tasks, since different VLMs or visual foundation models showcase unique characteristics. Probe3D~\cite{el2024probing} posits the 3D awareness of some visual foundation models, like DINOv2 present well for depth and surface normals. Lexicon3D~\cite{man2024lexicon3d} finds diffusion models benefit geometric tasks. 
Nonetheless, there is still a lack of studies on how to aggregate these heterogeneous foundation models into the 3D model which naturally excels at geometric knowledge extraction and locating 3D spatial objects.

To explore various foundation model properties when fusing multi-view posed image features to 3D space and enabling model distillation, we first conduct a pilot study to analyze and compare the heterogeneous results shown in Fig.~\ref{fig:motivation_fig}. We can observe that the distribution in terms of the feature embeddings from each 2D foundation model mainly follows a gaussian-like feature distribution. At the same time, when clustering these features into specific clusters, the results illustrate heterogeneous and complementary effects across different foundation models. This is also due to the inconsistency across different posed images when the 2D model encounters complex image contexts. This motivates us to develop a new method to harmonize this heterogeneous knowledge into a 3D model and handle such noisy inconsistency from the fused 2D feature embeddings.
\begin{table*}
    \centering
    \caption{\textbf{Comparison of Vision Foundation Models.} Although all utilize the same Vision Transformer (ViT) backbone, they greatly differ in their training paradigms, including data, image resolutions, and training objectives, which lead to diverse representation biases.}
    \vspace{-2mm}
    \label{tab:vfm}
    \resizebox*{0.8\linewidth}{!}{
    \begin{tabular}{c|cccc}
    \hline
    Model & Training Dataset & Dataset Size & Architecture & Objective \\
    \hline
    ViT~\citep{dosovitskiy2020image} & ImageNet-1k/21k & 1.2M/14.2M & ViT-B/L/G & Supervised classification \\
    DINOv2~\citep{oquab2023dinov2} & LVD-142M & 142M & ViT-L/14 & Discriminative self-supervised learning \\
    CLIP~\citep{radford2021learning} & WebImageText & 400M & ViT-L/14& Image-text contrastive learning \\
    Stable Diffusion~\citep{rombach2022high} & LAION & 5B & UNet & Image-Text/Image Generation \\
    \hline
    \end{tabular}}
    \vspace{-5mm}
\end{table*}

In this paper, we present \textit{\textbf{C}}ross-modal and \textit{\textbf{U}}ncertainty-aware \textit{\textbf{A}}gglomeration for \textit{\textbf{O}}pen-vocabulary \textit{\textbf{3D}} Scene Understanding, dubbed \textbf{CUA-O3D}, the first method to integrate multiple 2D foundation models into one 3D model for scene understanding. Deterministic uncertainty estimation is further introduced to adaptively distill and harmonize the heterogeneous 2D feature embeddings from these models. 
We show that there are potential inconsistencies yet complementarity when attempting to distill different foundation models. Based on our pilot study, we first propose to leverage distillation loss to supervise 3D model training given several available 2D feature embeddings. Our 3D model consists of independent projection layers to be mapped with one VLM or visual foundation model under feature supervision which helps with reconciling the entanglements from heterogeneous distributions. To resolve the potential noises from 2D models which mainly come from insufficient contexts, a novel deterministic uncertainty estimation is tailored to adaptively weight the knowledge distillation which can be modeled like a gaussian likelihood that follows the distributions as shown at the top of Fig.~\ref{fig:motivation_fig}. Specifically, regarding each projection layer to be mapped with the specific 2D model, we tailor an observation noise scalar prediction independently to capture how much the noise is contained in feature supervision during training, which is termed uncertainty-aware learning. Moving one step forward, as can be observed the distribution in terms of Stable Diffusion shows a bit shift away from the center scale, and usually carries broader value ranges and heavy-tailed (``spike'') values in the projected feature embeddings. A de-mean operation is then adopted to re-center the feature scales, being able to reduce the impact from anomaly points while still allowing points with small scale to guide the 3D model training.

As we show experimentally on ScannetV2~\cite{dai2017scannet} and Matterport3D~\cite{chang2017matterport3d}, our approach allows the 3D model agglomerates heterogeneous knowledge and reconciles with potential noises from various 2D feature supervisions. Extensive experiments demonstrate that our method not only advances open-vocabulary segmentation but also achieves competitive cross-domain alignments and spatial perception capabilities. Additionally, we also validate that our method can achieve significant downstream performance after distillation. In summary, the contributions of this work are:

\begin{itemize}
    \item To the best of our knowledge, we are the first one to investigate the agglomeration of cross-modal knowledge distillation from 2D models to a 3D model, given various strong foundation models available. 
    
    \item We analyze the heterogeneous yet complementary feature embeddings from multiple 2D models and incorporate both semantic- and geometric-aware knowledge into one single 3D model.
    
    \item We further propose a deterministic uncertainty estimation to enable the 3D model predict independent observation scalar to capture the noise and resolve the heterogeneity from various feature supervisions.

    \item We evaluate our method in a wide set of experiments from 3D open-vocabulary segmentation and present competitive cross-domain validation of our method, while also demonstrating strong downstream performances after distillation.
\end{itemize}

\section{Related Works}
\label{sec:relate}

\noindent\textbf{Open-Vocabulary (OV) 3D scene understanding} advances over the previous large corpus of close-set approaches \cite{chen2017deeplab,long2015fully,ronneberger2015u,badrinarayanan2017segnet,zhao2017pyramid,liu2024less,xu20243d}, allowing robust zero-shot reasoning and alleviating the need for annotations. 
Recent advances in Visual-Language Models (VLMs)~\cite{radford2021learning, jia2021scaling} have driven OV models towards remarkable levels of robustness with numerous emerging approaches tackling OV in image semantic segmentation~\cite{ghiasi2022scaling, li2022languagedriven, liang2023open, xu2023side}, object detection~\cite{zhou2022detecting, bangalath2022bridging}, and recently universal segmentation~\cite{zhang2023simple}.
Differently, OV for 3D scene understanding (OV3D) is limited in the data availability for training a \textit{purely fundamental} 3D VLM.
Alternatively, the community achieves OV3D by distilling zero-shot knowledge from recent VLMs~\cite{radford2021learning, li2022languagedriven} and by mapping point cloud features to a queryable CLIP space.
In 3D semantic segmentation, ConceptFusion~\cite{conceptfusion} fuses VLM representations from multiple views into 3D points.
Some methods~\cite{takmaz2023openmask3d} extends OV to 3D instance semantic segmentation 
based on CLIP~\cite{radford2021learning} or SAM~\cite{Kirillov_2023_ICCV} to align the 3D space with language space while forcing instance-mask constraint.
However, the recent OV3D methods heavily rely on the zero-shot knowledge of the 2D VLMs without investigating the reliability of the projected 2D feature representation.
In this work, we move forward and explore how to aggregate knowledge from various 2D foundation models.



\vspace{5pt}
\noindent\textbf{Knowledge distillation} (KD)~\cite{hinton2015distilling,park2019relational} aims at training compact student models with the supervision from more powerful and larger teacher models~\cite{li2022blip}. 
Introduced by the first work~\cite{hinton2015distilling}, the student model is trained to mimic the prediction behavior of the teacher model and has been extensively explored in subsequent works~\cite{romero2014fitnets,yim2017gift,ahn2019variational,zhao2022decoupled,li2021reskd,zhao2020mgsvf}, which has been applied successfully in a wide range of tasks going from supervised-training~\cite{mirzadeh2020improved,wu2021peer,touvron2021training}, to network compression~\cite{sau2016deep,chen2017learning,polino2018model} and to domain adaptation~\cite{zhao2020multi,hou2021visualizing,he2019knowledge}. 
Recently, following the explosion of Visual-Language Models like CLIP~\cite{radford2021learning} and ALIGN~\cite{jia2021scaling}, KD has been introduced to efficiently transfer knowledge between different modalities~\cite{li2022blip,ma2022open,gu2021open}, and recently to bridge the gap between the text and 3D point cloud modalities~\cite{chen2023clip2scene,yao20223d,huang2023clip2point,zhang2022pointclip,peng2023openscene,ding2022language,yang2023regionplc}. Recently, AM-RADIO~\cite{ranzinger2024radio} describes a general methodology for distilling multiple distinct foundation models into one, but still focus on only 2D domain.
%
%
We primarily focus on studying and tackling the ambiguity of the distilled representations between image and point cloud modalities.

\vspace{5pt}
\noindent\textbf{Uncertainty estimation}
 has been widely investigated in various tasks~\cite{kendall2017uncertainties, hu2020uncertainty, lakshminarayanan2017simple, laves2020well, oh2018modeling, guo2022uncertainty,mukhoti2023deep,kendall2018multi} which is capable of addressing the problem of quantifying the uncertainty of predictions by model. Uncertainty Estimation can be broadly classified into: (i) aleatoric estimation~\cite{kendall2017uncertainties, wang2021bayesian, nix1994estimating} that usually dues to the underlying uncertainty in the measurement which utilizes the extra network to be trained from scratch to approximate a heteroscedastic distribution by maximizing the likelihood of the system, and (ii) epistemic estimation~\cite{graves2011practical, lakshminarayanan2017simple, gal2016dropout} that induces the uncertainty by the model parameters in low-data regimes as parameter estimation becomes noisy, respectively. 
In 3D point clouds, uncertainty estimation finds applications in incremental learning~\cite{yang2023geometry}, semantic segmentation~\cite{cortinhal2020salsanext}, and domain adaptation~\cite{saltori2022gipso, wang2020train, saltori2023compositional}.
Unlike previous works, we seek a deterministic uncertainty estimation for supervising 3D model training under uncertainty awareness for the ambiguity between 2D image and 3D point modalities.


\begin{figure}[t]
  \centering
  \includegraphics[width=0.46\textwidth]{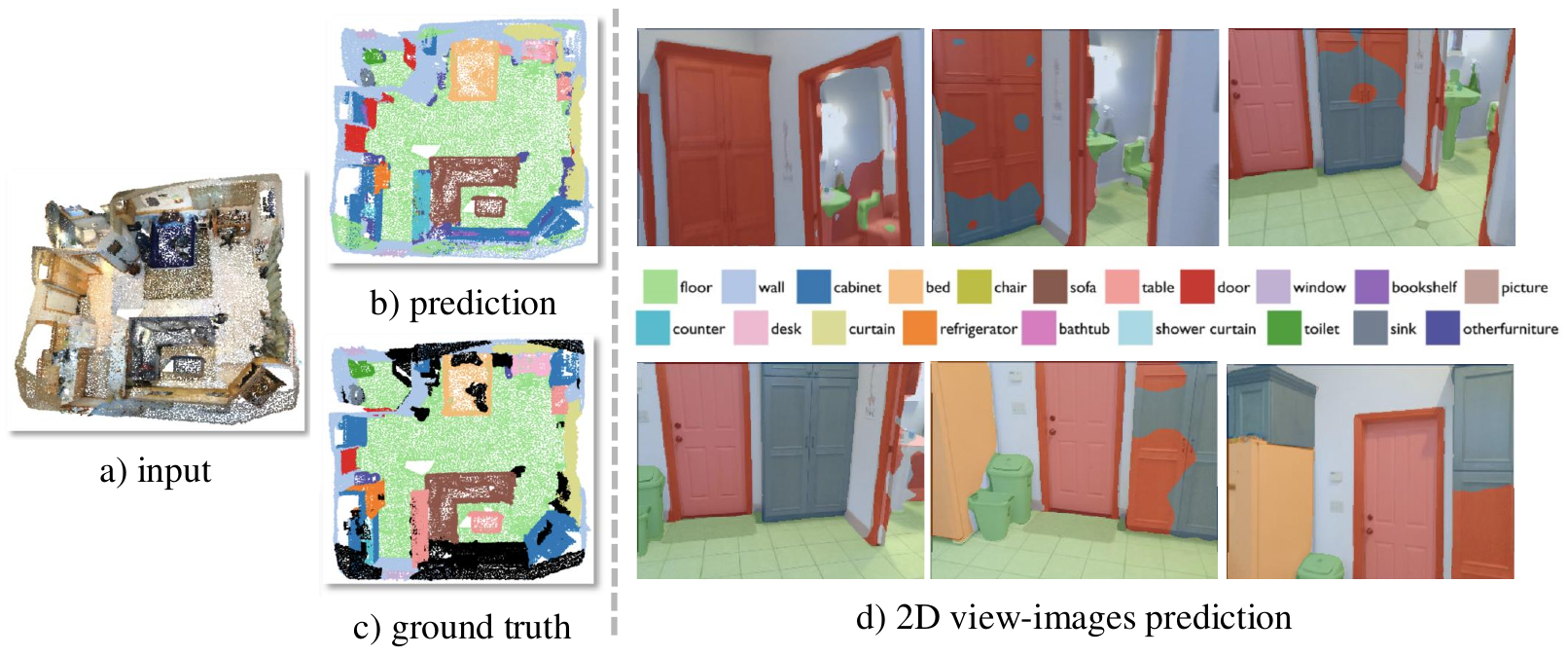}
  \vspace{-0.2cm}
  \caption{Preliminary study on image embedding ambiguity. VLM embeddings show inconsistent segmentations across multi-view images (\textit{e.g.} \textcolor{cabinet_color}{\textbf{\textit{cabinet}}}). The guidance with ambiguous embeddings may be detrimental for supervising a 3D model training.
  }
  \label{fig:pilot}
  \vspace{-0.6cm}
\end{figure}

\section{Methodology}
\label{sec:method}

In this section, we first describe the preliminary open-vocabulary 3D scene understanding task in Sec.~\ref{sec:ov3d}. Then we elucidate inconsistent results across multi-view posed images to demonstrate the necessity of uncertainty-aware training to alleviate such issues and embrace various 2D foundation models in Sec.~\ref{sec:incon}. Our Cross-Modal and Uncertainty-Aware Agglomeration (\textbf{CUA-O3D}) method will be depicted in Sec.~\ref{sec:crossAgg}, including Distillation agglomeration and Deterministic uncertainty estimation.

\subsection{Open-Vocabulary 3D Scene Understanding} 
\label{sec:ov3d}

In standard 3D semantic segmentation, the training set $\mathcal{T}_{train}$ includes point clouds and dense point-level annotations. Each point cloud $\mathcal{X} = { p_i \in \mathbb{R}^3, i \in [0, N-1]}$, consists of $N$ points $p_i$ with corresponding point-level annotations $\mathcal{Y}$. These annotations are assigned from a predefined set of class indices $\mathcal{K} = [1, \dots, K]$, where each index corresponds to a specific class name in the vocabulary $\mathcal{V} = [v_1, \dots, v_k]$. Given $\mathcal{T}_{train}$ and $\mathcal{K}$, the objective is to learn a deep neural network correctly assigning a label from $\mathcal{K}$ to each point $p_i \in \mathcal{X}$. 
In contrast, open-vocabulary 3D semantic segmentation (OV3D) aims to segment $\mathcal{X}$ using an arbitrary vocabulary $\mathcal{V}$. 
Recent approaches achieve this with a pre-trained VLM~\cite{peng2023openscene,takmaz2023openmask3d}.
The VLM provides this common embedding space through two distinct encoders, namely, a vision encoder $f^{2D}: \mathcal{I} \rightarrow \mathbf{Z}$ and a text encoder $f^{txt}: \mathcal{V} \rightarrow \mathbf{Z}$. The common practice is to train a 3D vision encoder $\theta_{3D}$ to align to $f^{2D}$ embeddings, bridging the modality gap and defining $\theta_{3D}: \mathcal{X} \rightarrow \mathbf{Z}$. After training, $f^{txt}$ encodes $\mathcal{V}$, and class predictions are computed via similarity matching between 3D point cloud and textual vocabulary embeddings.

\subsection{Preliminary Observation}
\label{sec:incon}

We first conduct a simple qualitative study to analyze how embedding ambiguity affects $f^{2D}$ predictions before and after projection to the point cloud space based on commonly used Lseg~\cite{li2022languagedriven}. We analyze the consistency of the predictions for the same object appearing in multi-view images.
Given a pre-trained vision-language model $f^{2D}$ and multi-view images $\mathcal{I}$, we query $f^{2D}$ with known vocabularies $\mathcal{V}$ and report the qualitative results over multiple ScanNetV2~\cite{dai2017scannet} view-images in Fig.~\ref{fig:pilot}. We notice that predictions are consistent for the classes \textcolor{wall_color}{\textbf{\textit{wall}}}, \textcolor{floor_color}{\textbf{\textit{floor}}}, and \textcolor{door_color}{\textit{\textbf{door}}} while showing inconsistency over the class \textcolor{cabinet_color}{\textbf{\textit{cabinet}}}.
After projection to the 3D point cloud space (\textit{left}), the projected prediction inherits this ambiguity, resulting in further detrimental 3D model distillation. Regarding the incorporation of various 2D foundation models shown in Fig.~\ref{fig:motivation_fig}, how to harmonize the heterogeneous characteristics also matters. 
This simple study highlights the need for a reliable uncertainty measure capturing embedding ambiguity that we hope to shed new light on the future works on this task.

\subsection{Cross-Modal Agglomeration}
\label{sec:crossAgg}

\begin{figure*}[t]
    \centering
    \includegraphics[width=0.9\textwidth]{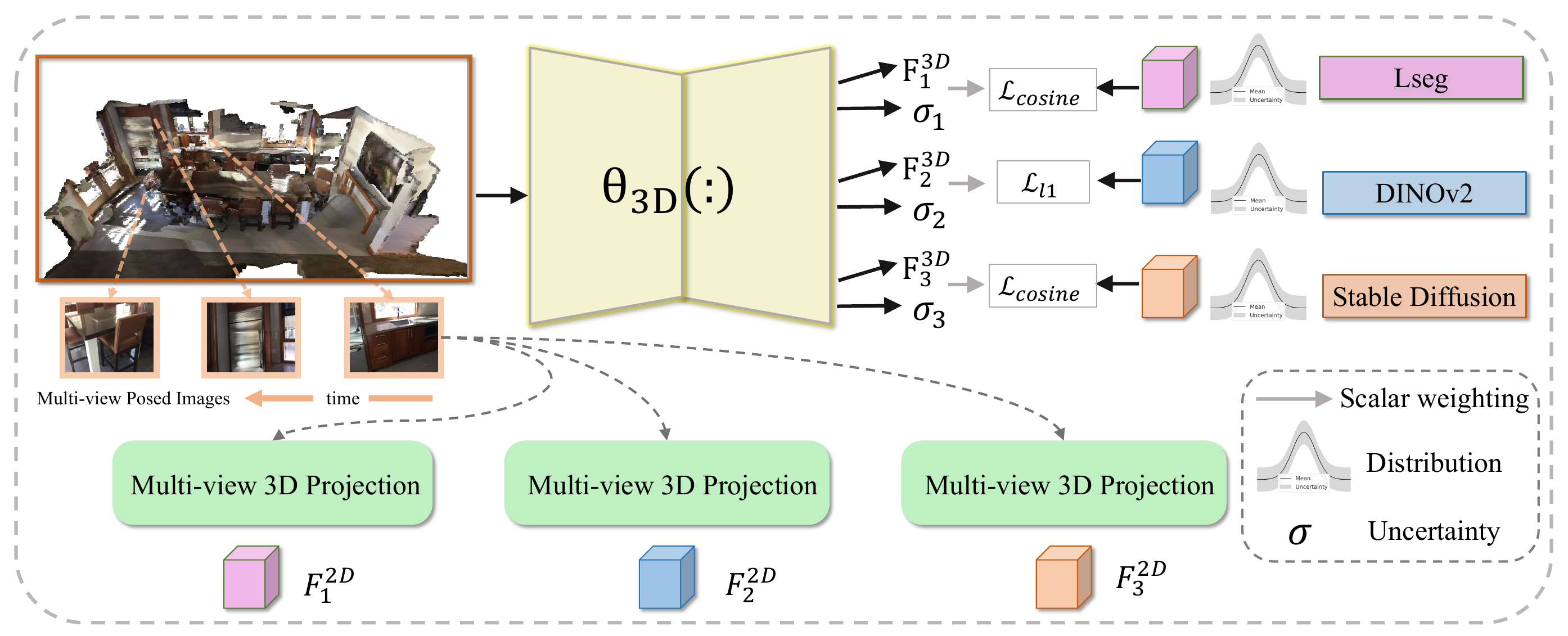}
    \vspace{-0.2cm}
    \caption{Overview of CUA-O3D. We first utilize Lseg, DINOv2 and Stable Diffusion model to extract multi-view posed image embeddings and then use multi-view 3D projection to obtain the projected 3D features $F^{2D}_{i}$ to supervise the 3D model training. Three MLP layers are established to map with each 2D model supervisions independently, while a specific noisy scalar prediction $\sigma_{i}$ through a deterministic uncertainty estimation will be learned and adopted to adaptively weight the corresponding distillation loss $\mathcal{L}$.}
\label{fig:method}
\vspace{-0.4cm}
\end{figure*}

We provide an overview of our~\textbf{CUA-O3D} in Fig.~\ref{fig:method}. Our approach leverages a 3D encoder backbone $\theta_{3D}$ to transform the input point cloud $\mathcal{X}$ into a 3D sparse point cloud features $F^{3D}$. Concurrently, we use several pre-trained vision encoders $f^{2D}_{i}$, such as CLIP, DINOv2, and Stable Diffusion, to map multi-view images $\mathcal{I}$ to dense image features separately, which are then projected to yield sparse image features $F^{2D}_{i}$. Then, we construct three projection layers with a simple MLP to map $F^{3D}_{i}$ with each 2D model through the corresponding distillation loss. Meanwhile, the 3D model is designed to output independently deterministic uncertainty-aware observation scalar $\sigma_{i}$ to adaptively weigh the feature supervisions. After training, we use $\theta_{3D}$ for the main task of open-vocabulary 3D semantic segmentation via matching with text embeddings $F^{txt}$~\cite{takmaz2023openmask3d} and drop the uncertainty module.

\begin{sloppypar}
\vspace{5pt}


\noindent\textbf{Distillation agglomeration.}
The distillation phase aims to align the point embeddings $F^{3D}$ to image embeddings $F^{2D}$ obtained from each frozen pre-trained visual encoder $f^{2D}_{i}$. This is a common practice in recent OV3D approaches~\cite{takmaz2023openmask3d}, which leads to a shared embedding space between image, text, and point cloud modalities.
Given a point cloud $\mathcal{X}$ paired with multi-view posed images $\mathcal{I}$, we learn 3D sparse features $F^{3D}$ by employing the well-established MinkowskiNet~\cite{choy20194d} as a 3D sparse convolutional encoder $\theta_{3D}$. The encoder $\theta_{3D}$ outputs a sparse set of point-wise feature vectors $F^{3D} = \theta_{3D}(\mathcal{X})$, where each feature is associated with an input point $p_i \in \mathcal{X}$. 
%
%
Similarly, we distill the multi-view image features from $\mathcal{I}$ using each frozen vision encoder $f^{2D}_{i}$ separately, where $\{ f^{2D}_{i} \in f^{2D} | f^{2D}_{Lseg}, f^{2D}_{DINOv2}, f^{2D}_{SD} \}$.
The output of $f^{2D}$ is a set of feature vectors $F^{2D}_{i} = f^{2D}_{i}(\mathcal{I})$ where each feature vector in $F^{2D}$ is associated to an input pixel $u$. After projection in the homogeneous coordinates space, we redefine $F^{2D}_{i}$ as a set of point-wise image features, corresponding to each $f^{2D}_{i}$.
We define the matches between each point $p$ and pixel $u$ by using the corresponding homogeneous coordinates $\tilde{p}$ and $\tilde{u}$, respectively. Once matches are established, we enforce the alignment between the predicted $F^{3D}_{i}$ and $F^{2D}_{i}$. 
Following \cite{peng2023openscene} and training $\theta_{3D}$ to minimize the distillation loss $\mathcal{L}_{distill}$, our independent distillation losses are defined as,
\end{sloppypar}

\vspace{-0.6cm}
\begin{equation}
\begin{split}
    & \text{Lseg,} \quad \mathcal{L}_{cos\_lseg} = 1 - \frac{F^{3D}_{1} \cdot F^{2D}_{1}}{\Vert F^{3D}_{1} \Vert_2 \cdot \Vert F^{2D}_{1} \Vert_2}
\label{eq:lseg_distillation}
\end{split}
\end{equation}

\vspace{-0.6cm}
\begin{equation}
\begin{split}
    & \text{DINOv2,} \quad \mathcal{L}_{l1} = \frac{1}{n} \sum_{i=1}^{n} |F^{3D}_{2} - F^{2D}_{2}|
\label{eq:dinov2_distillation}
\end{split}
\end{equation}

\vspace{-0.6cm}
\begin{equation}
\begin{split}
    \text{StableDiffusion,} \quad \mathcal{L}_{cos\_sd} = 1 - \frac{F^{3D}_{3} \cdot F^{2D}_{3}}{\Vert F^{3D}_{3} \Vert_2 \cdot \Vert F^{2D}_{3} \Vert_2}
\label{eq:sd_distillation}
\end{split}
\end{equation}
which corresponds to minimizing the distance between each $F^{3D}_{i}$ and $F^{2D}_{i}$, and leading to final distillation loss as,
\begin{equation}
  \mathcal{L}_{distill} = \mathcal{L}_{cos\_lseg} + \mathcal{L}_{l1} + \mathcal{L}_{cos\_sd},
\end{equation}
where we will study the distillation loss choice in our supplementary materials.

To further alleviate the impact from Stable Diffusion $F^{2D}_{3}$ which contains sharp values in the projected feature embeddings, we then adopt a de-mean operation to re-center the feature scales, reducing the impact from anomaly points while still allowing points with small scale to guide the 3D model training:

\vspace{-0.3cm}
\begin{equation}
    F^{2D}_{3} = F^{2D}_{3} - \mu_{F^{2D}_{3}}
\end{equation}
where $\mu_{F^{2D}_{3}}$ is the mean of $F^{2D}_{3}$ along channel dimension.

\noindent\textbf{Deterministic uncertainty estimation.}
As we analyzed before that different 2D foundation models encapsulate unique characteristics and one single 2D model induces inherent inconsistency from multi-view posed image which necessitates appropriate measures to tackle. We then propose a simple yet effective deterministic uncertainty-aware observation scalar prediction to quantify embedding ambiguity within each cross-modal distillation, that learns the adaptive weights of various 2D feature supervisions under the cross-modal training.

Specifically, we devise the 3D model $\theta_{3D}$ with three independent noise scalar predictions $\sigma_{i}$ ~\textit{w.r.t} each 2D model $f^{2D}_{i}$. The output from the probabilistic model with weight $\mathrm{W}$ being analogous to the regression task can be modeled as gaussian likelihood as:

\vspace{-0.3cm}
\begin{equation}
    p(y | f^\mathrm{W}(x)) = N(f^\mathrm{W}(x), \sigma^2),
\end{equation}
here we assume the 2D foundation models follow similar modeling as demonstrated in Fig.~\ref{fig:motivation_fig}, and our dense alignment training can be regarded as continuous model output. Combining all three 2D models we used, we then define the multiple model outputs as:

\vspace{-0.3cm}
\begin{equation}
    p(y_1, ..., y_K | f^\mathrm{W}(x)) = \prod \limits_{i=1}^K p(y_i | f^\mathrm{W}(x)),
\end{equation}
\vspace{-0.3cm}

where index $i$ corresponds to the mapping for each 2D model, $\{f^{2D}_{Lseg}, f^{2D}_{DINOv2}, f^{2D}_{SD} \}$. Based on the maximum likelihood modeling and taking the regression task as an example, the log-likelihood of the model can then be optimized,

\vspace{-0.4cm}
\begin{equation}
    \log p(y | f^\mathrm{W}(x)) \propto 
-\frac{1}{2 \sigma^2}||y - f^\mathrm{W}(x)||^2 - \log \sigma,
\end{equation}
\vspace{-0.1cm}
where $\sigma$ denotes the model's observation noise scalar, being responsible for capturing the inherent uncertainty within 2D feature supervisions, which is mainly caused by heterogeneous and noisy feature embeddings from various 2D foundation models. Assuming that we have three outputs corresponding to Lseg, DINOv2, and Stable Diffusion, each following gaussian-like distribution, we then have:

\vspace{-0.6cm}
\begin{equation}
\begin{split}
    p(y_1, y_2, y_3 | f^\mathrm{W}(x)) & =
    \prod \limits_{i=1}^3 p(y_i | f^\mathrm{W}(x)) \\
    & = \prod \limits_{i=1}^3 N(y_i; f^\mathrm{W}(x), \sigma_i^2),
\end{split}
\end{equation}
\vspace{-0.1cm}
and then, we formulate our training objective $\mathcal{L}_{distill}(\mathrm{W}, \sigma_1, \sigma_2, \sigma_3)$ for multiple mappings:

\vspace{-0.4cm}
\begin{equation}
\vspace{-0.2cm}
\begin{split}
     & \mathcal{L}_{distill} = 
    - \log p(y_1, y_2, y_3 | f^\mathrm{W}(x)) \\
    &\propto
    \frac{1}{2 \sigma_1^2} \mathcal{L}_{cos\_lseg}
    + \frac{1}{2 \sigma_2^2} \mathcal{L}_{l1} 
    + \frac{1}{2 \sigma_3^2} \mathcal{L}_{cos\_sd} 
    + \log \sigma_1 \sigma_2 \sigma_3,
\end{split}
\label{eq:loglihood}
\end{equation}
which leads to our overall training objective: $\mathcal{L} = \mathcal{L}_{distill}$.

To the best of our knowledge, this is the first work to explore deterministic uncertainty-aware modeling for agglomerating multiple 2D foundation models into a single unified 3D model, aiming toward the development of a potential foundational 3D model. In the last term of Eq.~\ref{eq:loglihood}, each supervision signal from a 2D model contributes to learning adaptive weighting during training. Specifically, the uncertainty parameter $\sigma_i$, which characterizes the noise level associated with the $i$-th 2D model $f^{2D}_i$, dynamically adjusts the contribution of the corresponding loss term $\mathcal{L}_i$. As $\sigma_i$ increases—indicating higher uncertainty—the effective weight of $\mathcal{L}_i$ decreases, thereby down-weighting less reliable supervision. Meanwhile, each $\sigma_i$ is implicitly regularized to prevent excessive growth, ensuring that all supervision sources contribute meaningfully to the optimization. We visualize the evolution of $\sigma_i$ throughout training in the supplementary material. To further ensure stable optimization, we add a small constant (set to $1.0$) to each $\sigma_i$ to prevent the loss from becoming negative during training.

\vspace{-0.6cm}
\begin{equation}
\begin{split}
     \log \sigma_{i} \to \log( 1.0 + \sigma_{i}).
\end{split}
\vspace{-0.8cm}
\end{equation}


\section{Experiments}
\label{sec:exp}


We run extensive experiments over a wide set of tasks, going from open-vocabulary segmentation and cross-domain generalization to the evaluation with fine-grained class vocabularies.
This section is organized as follows.
Sec.~\ref{sec:datasets}-~\ref{sec:implementation} provide the dataset and implementation details of ~\textbf{CUA-O3D} used in our experiments.
Sec.~\ref{sec:qualitative_com} and Sec.~\ref{sec:open_voc_results} illustrate the potential when embracing various 2D foundation models and evaluate the open-vocabulary 3D semantic segmentation, compared with related methods.
Sec.~\ref{sec:cross_dataset_generalization} reports cross-domain generalization under common and fine-grained class evaluation while sec~\ref{sec:linear_probe} presents the downstream performance after distillation.
%
%
The final Sec.~\ref{sec:ablation} ablates and analyzes the improvements of each proposed components.

\subsection{Datasets}
\label{sec:datasets}

\noindent\textbf{ScanNetV2}~\cite{dai2017scannet, rozenberszki2022language} is a large-scale annotated indoor dataset. It includes over $2.5$ million camera views within more than $1.5k$ RGB-D scans, collected across $707$ diverse indoor environments such as offices and living rooms. The dataset is enriched with annotations, including 3D camera poses and point-level semantic segmentation. It is officially split into $1201$ training and $312$ validation scans sampled from $706$ different scenes, and $100$ scans test set with hidden ground truth. ScanNetV2 annotations cover $40$ semantic classes, while the official 3D semantic segmentation benchmark focuses on a subset of $20$ classes.

\noindent\textbf{Matterport3D}~\cite{chang2017matterport3d} is another large-scale RGB-D dataset collected for 3D scene understanding in indoor settings, containing $90$ buildings with multiple rooms on different floors captured using a Matterport Pro Camera. It provides $10.8k$ panoramic views within $90$ real, building-scale scenes processed from $194.4k$ RGB-D images. Each scene represents a residential building with multiple rooms and is annotated with camera poses and point-level semantic segmentation. The official 3D semantic segmentation benchmark evaluates performance across $21$ semantic classes.

\subsection{Implementation Details}
\label{sec:implementation}

We implement our method in the PyTorch framework~\cite{paszke2019pytorch} and employ experiments on a single NVIDIA A100 GPU for ScanNetV2 and Matterport3D, respectively.  We follow ~\cite{peng2023openscene} and use MinkUNet18A~\cite{choy20194d} as our 3D backbone starting from randomly initialized weights, and Lseg~\cite{li2022languagedriven} as our pre-trained VLM. During our training, we adopt Adam optimizer~\cite{kingma2014adam} with an initial learning rate of $1e-4$ and an exponential decay to train our pipeline for 50 epochs. For the teacher model parameter update, we set the momentum coefficient $\beta$ to $0.99$ and $\gamma$ to $1$. Besides, we employ the voxel size of $2$cm and batch size of $2$ for both the ScanNetV2 and Matterport3D experiments. Due to the GPU memory limitation, we uniformly sample $20k$ point features to train the model and input only the 3D point position without RGB information to the MinkowskiNet. We utilize random horizontal flip and elastic distortion as data augmentations over point clouds while following BP-Net~\cite{hu2021bidirectional} to apply color, jitter, and hue feature transformations over 2D feature embeddings.

\begin{table}[t]
    \centering
    \caption{Open-vocabulary 3D semantic segmentation results. We compare our CUA-O3D with recent fully supervised (\textit{Fully-sup.}) and zero-shot (\textit{Zero-shot}) baselines. Our method demonstrates competitive performance on both ScanNetV2 and Matterport3D. $^{\dagger}$ denotes results from origin paper based on Lseg.}
    \vspace{-0.2cm}
    \label{tab:results_segmentation}
    \tabcolsep 4pt
    \resizebox{1.0\columnwidth}{!}{%
    \begin{tabular}{l|l|cc|cc}
        \toprule
        \textbf{Type} & \textbf{Method} & \multicolumn{2}{c|}{\textbf{ScanNetV2}} & \multicolumn{2}{c}{\textbf{Matterport3D}} \\
        ~ & ~ & mIoU & mAcc & mIoU & mAcc \\
        \midrule

        \multirow{8}{*}{\textit{Fully-sup.}} & TangentConv~\cite{tatarchenko2018tangent} & 40.9 & - & - & 46.8 \\
        ~ & TextureNet~\cite{huang2019texturenet} & 54.8 & - & - & 63.0 \\
        ~ & ScanComplete~\cite{dai2018scancomplete} & 56.6 & - & - & 44.9 \\
        ~ & DCM-Net~\cite{schult2020dualconvmesh} & 65.8 & - & - & 66.2 \\
        ~ & Mix3D~\cite{nekrasov2021mix3d} & 73.6 & - & - & - \\
        ~ & SupCon~\cite{zheng2021weakly} & 69.2 & 77.7 & 53.1 & 63.4 \\
        ~ & LGround~\cite{rozenberszki2022language} & 73.2 & - & - & 67.2 \\
        ~ & MinkowskiNet~\cite{choy20194d} & 69.2 & 77.7 & 53.1 & 63.4 \\
        \midrule
        \multirow{1}{*}{\textit{Upper-bound}} & MinkowskiNet$^{reimple}$~\cite{choy20194d} & 68.96 & 77.41 & 54.12 & 65.57 \\
        \midrule
        \multirow{10}{*}{\textit{Zero-shot}} & MSeg Voting~\cite{lambert2020mseg}  & 45.6 & 54.4 & 33.4 & -  \\
        ~ & PLA~\cite{ding2023pla}  & 17.7 & 33.5 & - & -  \\
        ~ & CLIP2Scene~\cite{chen2023clip2scene}  & 25.1 & - & - & -  \\
        ~ & CNS~\cite{chen2024towards}  & 26.8 & - & - & -  \\
        ~ & CLIP-FO3D~\cite{zhang2023clip}  & 30.2 & 49.1 & - & -  \\
        ~ & RegionPLC~\cite{yang2023regionplc}  & 43.8 & 65.6 & - & -  \\
        ~ & DMA-text only~\cite{li2024dense}  & 50.5 & 63.7 & 39.8 & 49.5  \\
        ~ & \textcolor{lightgray}{OpenScene-3D}$^{\dagger}$~\cite{peng2023openscene}  & \textcolor{lightgray}{52.9} & \textcolor{lightgray}{63.2} & \textcolor{lightgray}{41.9} & \textcolor{lightgray}{51.2}  \\
        ~ & \textcolor{lightgray}{OpenScene-2D3D}$^{\dagger}$~\cite{peng2023openscene}  & \textcolor{lightgray}{54.2} & \textcolor{lightgray}{66.6} & \textcolor{lightgray}{43.4} & \textcolor{lightgray}{53.5}  \\
        ~ & OpenScene$^{reimple}$-3D~\cite{peng2023openscene}  & 51.6 & 63.1 & 40.5 & 48.8  \\
        ~ & OpenScene$^{reimple}$-2D3D~\cite{peng2023openscene}  & 52.2 & 65.4 & 41.5 & 50.6  \\
       \rowcolor{tabhighlight} ~ & (\textit{Ours}) CUA-O3D (3D) & 54.1 & 64.1 & 41.3 & 49.5  \\
       \rowcolor{tabhighlight} ~ & (\textit{Ours}) CUA-O3D (2D3D) & \textbf{55.3} & 65.6 & \textbf{42.2} & \textbf{50.9}  \\
        
        \bottomrule
    \end{tabular}
    }
    \label{tab:sota_ov_seg}
    \vspace{-0.5cm}
\end{table}

\begin{figure*}[tb]
  \centering
  \includegraphics[width=0.95\textwidth]{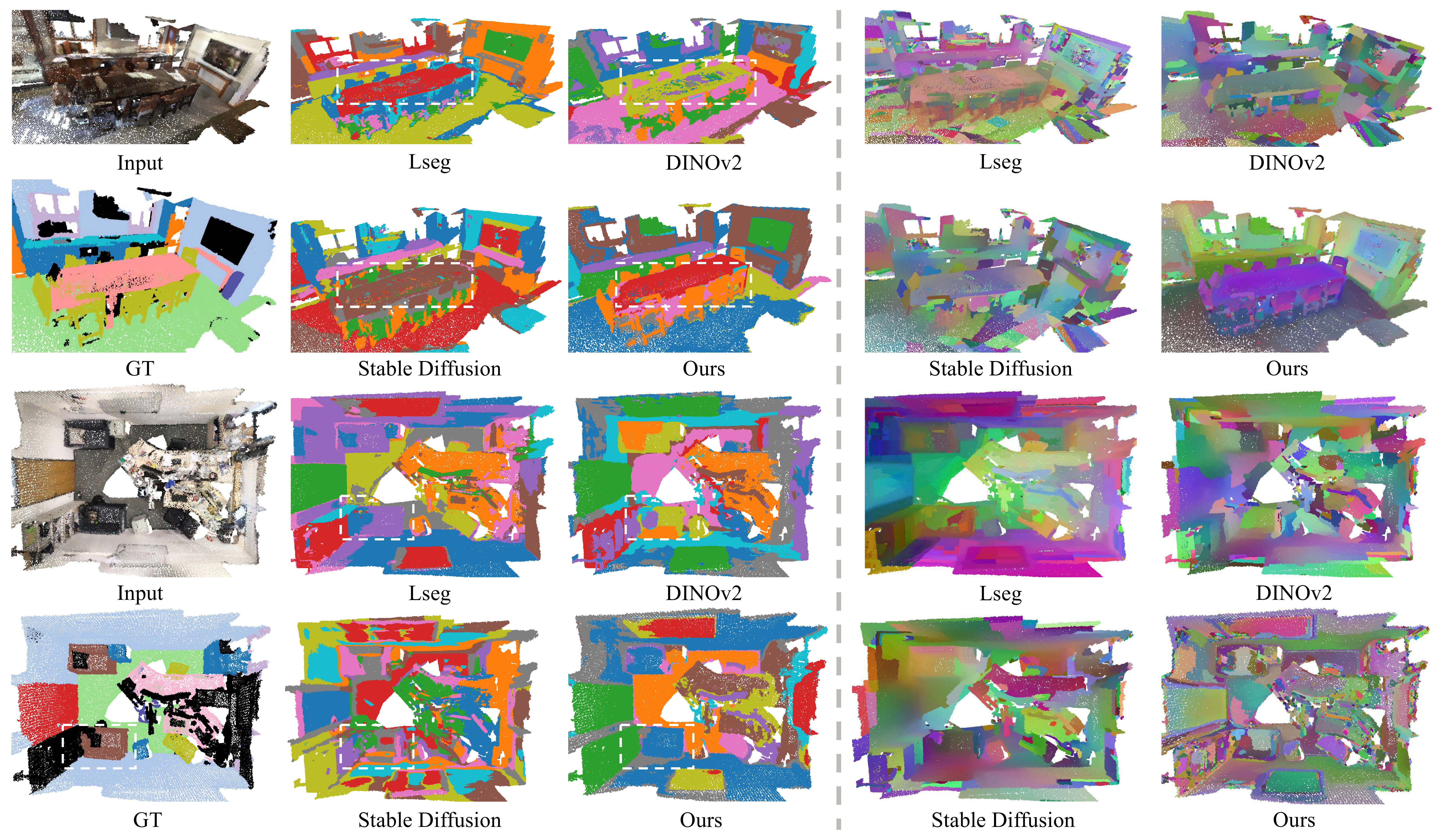}
  \vspace{-0.3cm}
  \caption{
      Left side: $K\_Means$ is tapped to cluster the projected 3D feature embeddings based on Lseg. DINOv2, Stable Diffusion and our final distilled feauture predicted by the 3D model. Right side: UMAP~\cite{SMG2020} is applied to project high-dimension feature into low-dimension one to visualize the structural characteristics. White rectangle highlights the apparent heterogeneous yet complementary results.
  }
  \label{fig:qualitatives}
  \vspace{-0.5cm}
\end{figure*}

\subsection{Qualitative Comparisons}
\label{sec:qualitative_com}

To demonstrate the motivation of various 2D foundation models agglomeration, we investigate the results using \textit{K-Means} to cluster the projected feature embeddings from different 2D models and our distilled features as shown in Fig.~\ref{fig:qualitatives}. As can be observed that different 2D model performs heterogeneous yet complementary results. Likewise, the \textit{table} from the top-left sample in Fig.~\ref{fig:qualitatives} displays various clustering results, leading to our agglomerated model being able to output more accurate and consistent clusters. Meanwhile, we also utilize UMAP~\cite{SMG2020} to better illustrate the intrinsic characteristics when adopting a specific 2D model to employ the 3D model distillation, since the output feature embeddings from various dimensions of DINOv2 and Stable Diffusion are not elaborated for direct matching with text embeddings. As shown on the right side of Fig.~\ref{fig:qualitatives}, DINOv2 indicates smoother and more consistent results though Lseg has been trained to align with the text encoder before within dense supervision. Stable Difusion presents intriguing geometric characteristics, all of which are capable of agglomerating potential foundation 3D models using distillation.

\subsection{Open-Vocabulary 3D Semantic Segmentation}
\label{sec:open_voc_results}

\begin{figure}[ht]
  \centering
  \includegraphics[width=0.48\textwidth]{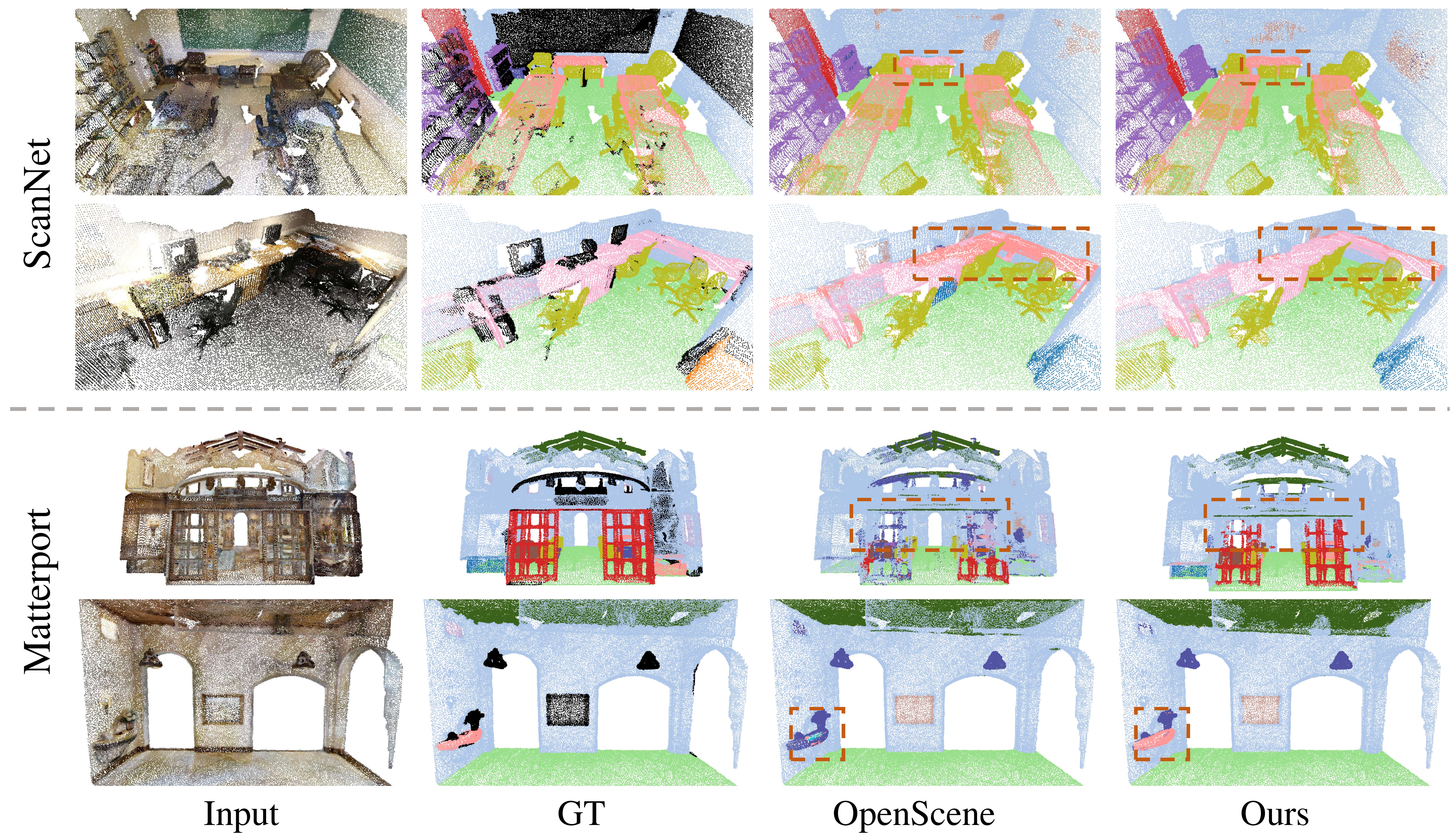}
  \vspace{-0.7cm}
  \caption{
        Open-vocabulary 3D semantic segmentation comparisons in terms of ScanNetV2 and Matterport3D. Our approach displays superior performance over the OpenScene, which is regarded as our \textit{baseline}. Best view \textit{zoom in and out}.
  }
  \label{fig:seg_vis}
  \vspace{-0.3cm}
\end{figure}

To showcase that our method CUA-O3D can boost the performances of the open-vocabulary 3D semantic segmentation model, we compare our method with existing works within two types of settings, including fully supervised (\textit{Fully-sup.}) and zero-shot. From Table~\ref{tab:sota_ov_seg}, our method surpasses the recent work OpenScene$^{reimple}$~\cite{peng2023openscene} with $+2.5\%$ mIoU under 3D-distill and $+3.1\%$ mIoU under 2D3D-ensemble on ScanNetV2 \textit{val set}, and $+0.8\%$ mIoU under 3D-distill and $+0.7\%$ mIoU under 2D3D-ensemble on Matterport3D \textit{val set}, respectively. This further proves our method not only distills heterogeneous yet complementary knowledge into the 3D model but also reconciles with 2D noisy supervisions. This is achieved by the proposed deterministic uncertainty estimation which adaptively captures inherent noise and then weights the corresponding distillation. Supervised by various 2D foundation models, like Lseg, DINOv2, and Stable Diffusion, the 3D model learns to align with the open-vocabulary features together with spatial and geometric awareness. Some open-vocabulary 3D semantic segmentation visualizations are shown in Fig.~\ref{fig:seg_vis}. Additional experiments with AMRADIO~\cite{ranzinger2024radio} can be referred to our supplementary material.

\begin{table}[t]
    \centering
    \caption{
    Cross-dataset evaluation. We evaluate the cross-dataset generalization capability of CUA-O3D. We perform this experiment when training on ScanNetV2 and evaluating on Matterport3D (ScanNetV2 $\rightarrow$ Matterport3D), and \textit{vice versa}.
    }
    \vspace{-0.2cm}
    \resizebox{.75\columnwidth}{!}{%
    \begin{tabular}{l|cc}
    \toprule
        \multicolumn{3}{c}{ScanNetV2 (\textit{train}) $\rightarrow$Matterport3D (\textit{eval})} \\
    \toprule
        \textbf{Method} & \textbf{mIoU} & \textbf{mAcc} \\
        \midrule
        OpenScene~\cite{peng2023openscene} & 36.0 & 48.0 \\
        (\textit{Ours}) CUA-O3D & \textbf{37.4} (\textcolor{red}{+1.4}) & \textbf{49.2}(\textcolor{red}{+1.2}) \\

        \toprule
        \multicolumn{3}{c}{Matterport3D (\textit{train})$\rightarrow$ScanNetV2 (\textit{eval})} \\
        \toprule
        OpenScene~\cite{peng2023openscene} & 36.5 & 44.0 \\
        (\textit{Ours}) CUA-O3D  & \textbf{38.6} (\textcolor{red}{+2.1}) & \textbf{46.6}(\textcolor{red}{+2.6}) \\
    \bottomrule
    \end{tabular}
    }
    \label{tab:cross_dataset_seg}
    \vspace{-0.5cm}
\end{table}

\subsection{Cross-Dataset Generalization}
\label{sec:cross_dataset_generalization}

We study the generalization capability of our CUA-O3D to unseen datasets without further fine-tuning (\ie, zero-shot).
This is achieved by evaluating the cross-dataset performance when training and evaluating different datasets.
We use ScanNetV2 and Matterport3D as training and evaluation datasets respectively, and analyze the cross-dataset results when training on ScanNetV2 and evaluating on Matterport3D, and \textit{vice versa}.
 Table~\ref{tab:cross_dataset_seg} and Table~\ref{tab:cross} report the cross-dataset results and with different granularities, respectively. 

\vspace{-0.3cm}

\begin{table*}[t]
    \centering
    \caption{Comparison on cross-dataset generalization. Both CUA-O3D and OpenScene
are trained on ScanNet, and zero-shot tested on the Matterport3D dataset. $\ddagger$ denotes the pure 3D results obtained from the official released model. K = 21 is
derived from the original Matterport3D benchmark, while K = 40, 80, 160, is 
K most common categories from the NYU label set provided in the benchmark.}
    \vspace{-0.2cm}
    \label{tab:results_segmentation_fine}
    \tabcolsep 4pt
    \resizebox{1.5\columnwidth}{!}{%
    \begin{tabular}{l|cc|cc|cc|cc}
        \toprule
         \textbf{Method} & \multicolumn{2}{c}{\textbf{Matterport21}} & \multicolumn{2}{c}{\textbf{Matterport40}} & \multicolumn{2}{c}{\textbf{Matterport80}} & \multicolumn{2}{c}{\textbf{Matterport160}} \\
         ~ & mIoU & mAcc & mIoU & mAcc & mIoU & mAcc & mIoU & mAcc \\
        \midrule
         OpenScene$^{\ddagger}$~\cite{peng2023openscene} & 36.0 & 48.0 & 21.1 & 27.5 & 10.8 & 13.9 & 6.0 & 8.1 \\
        (\textit{Ours}) CUA-O3D (2D3D) & \textbf{37.4} & \textbf{49.2} & \textbf{23.3} & \textbf{30.2} & \textbf{12.2} & \textbf{16.3} & \textbf{6.1} & \textbf{8.4}  \\
        \bottomrule
    \end{tabular}
    }
    \label{tab:cross}
    \vspace{-0.5cm}
\end{table*}



We notice that our method improves the cross-dataset performance in both directions. As reported in Table~\ref{tab:cross_dataset_seg},
CUA-O3D consistently outperforms OpenScene in both directions with $+1.4\%$ mIoU improvements on ScanNetV2 $\rightarrow$ Matterport3D and $+2.1\%$ mIoU improvements on Matterport3D $\rightarrow$ ScanNetV2.
Interestingly, we notice that our approach also provides consistent superiority across different granularities in Table~\ref{tab:cross}, ranging among K = 21, 40, 60, and 160 common categories in terms of zero-shot evaluation on ScanNetV2 $\rightarrow$ Matterport3D.

\begin{table}[t]
    \centering
    \caption{
        Experimental results on ScanNetV2 and Matterport3D in terms of \textit{val} on linear probing evaluation. Upperbound-full sup. denotes the fully-supervised upperbounding results while Baseline init. means initialize the model from our baseline model and then perform linear probing evaluation.
        }
        \vspace{-0.2cm}
    \tabcolsep 4pt
    \resizebox{.99\columnwidth}{!}{%
    \begin{tabular}{l|l|cc|ccc}
        \toprule
        Type & \textbf{Method} & \multicolumn{2}{c}{\textbf{ScanNetV2}} & \multicolumn{2}{c}{\textbf{Matterport3D}} \\
        ~ & ~ & mIoU & mAcc & mIoU & mAcc \\
        \midrule
        Upperbound-fully sup. & MinkowskiNet~\cite{choy20194d} & 68.9 & 77.4 & 54.1 & 65.5 & \\
        Baseline init. & MinkowskiNet~\cite{choy20194d} & 54.4 & 64.7 & 36.1 & 43.0 & \\
        \midrule

        Concat & 3-heads concat & 62.1 & 72.7 & 45.8 & 55.3 & \\
        \midrule
        Separate & 3-heads average  & 61.7 & 72.0 & 45.4 & 55.0 & \\
        \midrule
        \multirow{3}{*}{Single-head} & Lseg-head & 59.9 & 71.5 & - & - & \\
        ~ & DINOv2-head & 61.7 & 72.2 & - & - & \\
        ~ & StableDiffusion-head & 61.4 & 72.1 & - & - & \\
        
        \bottomrule
    \end{tabular}
    }
    \label{tab:linear_probe}
    \vspace{-0.6cm}
\end{table}

\subsection{Linear Probing}
\label{sec:linear_probe}

In this section, we exploit how the trained 3D model will perform after agglomerating various 2D models. We then conduct experiments that employ linear probe learning based on the 3D model after distilling from various 2D models. Specifically, we construct a simple MLP layer on top of the frozen 3D model backbone and train the linear layer only following the fully-supervised manner. As shown in Table~\ref{tab:linear_probe}, the method concatenates all three mapping layers corresponding to Lseg, DINOv2, and Stable Diffusion, and then maps the concatenated features to close set label spaces, we can see this way can achieve the best performances, 62.1\% mIoU and 45.8\% mIoU on ScanNetV2 and Matterport3D \textit{val} after tuned on \textit{train}. Note that this obtains 7.7\% mIoU improvement over the model initialized from our baseline model. We can also observe that simply mapping the DINOv2 layer from the 3D model attains very competitive segmentation performance while mapping the Lseg layer only which has been trained before to align with the text encoder realizes an inferior one. These further insights that we shall seek more suitable 2D model selections to help develop potential foundational 3D models, whereas DINOv2 presents strong
generalizability and flexibility, which is consistent with the observation~\cite{man2024lexicon3d,el2024probing}.

\subsection{Ablative Studies}
\label{sec:ablation}

In this section, we study the improvements from each proposed component. As shown in Table~\ref{tab:ablations}, we begin by gradually adding each component to the 3D model training and find that only combining with Lseg and DINOv2 leads to marginal open-vocabulary 3D semantic segmentation which can be conjectured that DINOv2 has not been aligned with language space before though it excels at spatial perception ability. Then, the performance is boosted by 1.3\% mIoU when introducing Stable Diffusion supervision, while further improved by 2.1\% mIoU and 1.9\% mAcc via our proposed deterministic estimation to help the model adaptively harmonize the heterogeneous knowledge from various 2D models. Interestingly, if we apply auto-weighting to enable the model to learn by itself, the model training falls into collapse, which we surmise it is due to the minimal optimization objective and the model quickly gets into a trivial solution. Overall, CUA-O3D can improve the 3D model only from 51.4\% mIoU and 62.3\% mAcc to 54.1\% mIoU and 64.1\% mAcc, which further demonstrates the effectiveness of our method.

\begin{table}[!tb]
\label{ablationDAP}
\caption{
    Ablation: Contribution of each component by gradually adding into the final training, based on \textit{zero-shot segmentation}.
}
\vspace{-0.2cm}
\centering
\setlength\tabcolsep{2.0pt}
\resizebox{.99\columnwidth}{!}{%
    \begin{tabular}{cccccc|cc}
    \toprule
    \makecell*[c]{Baseline$_{Lseg}$} & $+DINOv2$ & $+SD$  & $+Unc$ & $+AutoW$ & $+DeMean$ & mIoU $\uparrow$ & mAcc $\uparrow$\\
     \midrule
    \checkmark  & & & ~ & ~ & ~ & 51.4 & 62.3 \\ 
    \checkmark & \checkmark &   & ~ & ~ & ~ & 51.7 & 63.3 \\
    \checkmark &  & \checkmark  & ~ & ~ & ~ & 51.4 & 62.4 \\
    \checkmark  & \checkmark & \checkmark & ~ & ~ & ~ & 52.7 & 62.6 \\
    \checkmark & \checkmark & \checkmark &\checkmark &  ~ & ~ & 53.5 & 64.2  \\
    \checkmark & \checkmark &\checkmark  &  ~ & \checkmark & ~ & $NAN$ & $NAN$ \\
    \checkmark &  \checkmark &\checkmark  &\checkmark & ~ & \checkmark & 54.1 & 64.1 \\
    \bottomrule
    \end{tabular}
}
\label{tab:ablations}
\vspace{-0.4cm}
\end{table}

\section{Conclusions}
\label{sec:con}

In this paper, we first investigate the cross-modal agglomeration from various 2D foundation models into one 3D model, in pursuit of a potential foundational 3D model. To resolve the heterogeneous bias and inherent noise from 2D feature supervisions, we then propose a deterministic uncertainty estimation to capture 2D model-specific uncertainties across diverse semantic and geometric sensitivities, which is then leveraged to weight the corresponding distillation loss adaptively. In this way, the trained 3D model performs competitive open-vocabulary segmentation while achieving robust cross-domain alignment and strong spatial perception ability, which hopes to shed new light on the community.

\clearpage

\vspace{1mm}
\noindent \textbf{Acknowledgments}
This work was supported by the MUR PNRR project FAIR (PE00000013) funded by the NextGenerationEU and the EU Horizon project ELIAS (No. 101120237).
We acknowledge the CINECA award under the ISCRA initiative for the availability of high-performance computing resources and support.



{
    \small
    \bibliographystyle{ieeenat_fullname}
    \bibliography{main}
}



\maketitlesupplementary

\begin{abstract}
This supplementary material provides additional details and analysis to support the main paper, as follows:
\begin{itemize}
    \item In Sec.~\ref{sec:loss_choice}, we describe the choice of distillation losses for various 2D foundation models.
    \item In Sec.~\ref{sec:amradio_com}, we conduct more experiments in terms of combining both Lseg~\cite{li2022languagedriven} and OpenSeg~\cite{ghiasi2022scaling} to supervise the 3D model training.
    \item In Sec.~\ref{sec:lseg_openseg}, we illustrate more comparisons with AMRADIO~\cite{ranzinger2024radio}.
    \item In Sec.~\ref{sec:sigma_vis}, we provide parameters evolution in terms of the deterministic uncertainty estimation during the training.
    \item In Sec.~\ref{sec:moremore_vis}, we provide more visualizations.
    \item In Sec.~\ref{sec:limitation}, we describe the potential limitation and future improvements.
\end{itemize}

\end{abstract}

\section{Distillation Loss Choice}
\label{sec:loss_choice}

In this section, we conduct more ablative studies to explain the distillation loss combination in our main paper. We first ablate three losses, including cosine similarity loss, L1 loss, and MSE (L2) loss, to explore how each one influences the 3D distilled model's open-vocabulary performance over the baseline model. As shown in Table~\ref{tab:loss_ablations}, we can find that when supervising the 3D model training with Lseg that has been aligned with language spaces, cosine similarity loss strikes the best. In contrast, both L1 and L2 loss significantly deteriorates the open-vocabulary semantic segmentation performance. We surmise that L1 and L2 loss cannot force the 3D model to align with the language space, meaning degraded semantic information learning, however, it may help with learning geometric awareness.
Regarding the pe-training mechanisms in terms of Lseg, DINOv2, and Stable Diffusion 2D foundation models, we opt to equip the 3D model with cosine similarity loss for Lseg and Stable Diffusion supervision since they have been trained with text embeddings. Moreover, we conducted additional analyses to study the effect of the 3D distillation model when attempting to use L1 or L2 loss to supervise from the DINOv2. As shown in Table~\ref{tab:loss_ablations_v2}, we can observe that adopting L1 loss for DINOv2 achieves the best one where we deduce this is due to the similar feature scales with Lseg's and L2 loss will decrease the supervised signals for the 3D model from feature embeddings which contains scales lower than 1.0. Therefore, we choose to use cosine similarity loss for Lseg and Stable Diffusion while L1 loss for DINOv2 in our main paper.  

\begin{table}[!tb]
\label{ablationLoss}
\caption{
    Ablation: loss ablation about Cosine Similarity Loss, L1 Loss and MSE (L2) Loss.
}
\vspace{-0.2cm}
\centering
\setlength\tabcolsep{2.0pt}
\resizebox{.99\columnwidth}{!}{%
    \begin{tabular}{cccc|cc}
    \toprule
    \makecell*[c]{Baseline$_{Lseg}$} & $+CosineLoss2$ & $+L1Loss$  & $+L2Loss$ & mIoU $\uparrow$ & mAcc $\uparrow$\\
     \midrule
    \checkmark & \checkmark &   & ~ & \textbf{51.4} & \textbf{62.3} \\
    \checkmark  & ~ & \checkmark & ~ & 46.6 & 57.0 \\
    \checkmark & ~ & ~ &\checkmark & 48.4 & 61.7  \\
    \bottomrule
    \end{tabular}
}
\label{tab:loss_ablations}
\end{table}

\begin{table}[!tb]
\label{ablationDAP}
\caption{
    Ablation: loss ablation for DINOv2 supervision when attempting with Cosine Similarity Loss, L1 Loss or MSE (L2) Loss.
}
\vspace{-0.2cm}
\centering
\setlength\tabcolsep{2.0pt}
\resizebox{.99\columnwidth}{!}{%
    \begin{tabular}{ccccc|cc}
    \toprule
    \makecell*[c]{Baseline$_{Lseg}$} & $+SD$  & $+DINOv2-Cosine$ & $+DINOv2-L1$ & $+DINOv2-L2$ & mIoU $\uparrow$ & mAcc $\uparrow$\\
     \midrule
    \checkmark  & \checkmark & \checkmark & ~ & ~ & 51.4 & 62.3 \\ 
    \checkmark  & \checkmark  & ~ & \checkmark & ~ & \textbf{52.7} & \textbf{62.6} \\
    \checkmark   & \checkmark & ~ & ~ & \checkmark & 51.7 & 63.3 \\
    \bottomrule
    \end{tabular}
}
\label{tab:loss_ablations_v2}
\end{table}

\begin{table}[!tb]
\label{ablationDAP}
\caption{
    Ablation: additional studies when introducing AMRADIO distillation.
}
\vspace{-0.2cm}
\centering
\setlength\tabcolsep{2.0pt}
\resizebox{.99\columnwidth}{!}{%
    \begin{tabular}{ccccc|cc}
    \toprule
    \makecell*[c]{Baseline$_{Lseg}$} & $+DINOv2$ & $+SD$ & $+AMRADIO$ & $+Unc$ & mIoU $\uparrow$ & mAcc $\uparrow$\\
     \midrule
    \checkmark  & & & ~ & ~ & 51.4 & 62.3 \\ 
    \checkmark & \checkmark &   & ~ & ~ & 51.7 & 63.3 \\
    \checkmark  & \checkmark & \checkmark & ~ & ~ & 52.7 & 62.6 \\
    \checkmark  & \checkmark & \checkmark & ~ & \checkmark & 53.5 & 64.2 \\
    \checkmark & \checkmark & \checkmark &\checkmark &  ~ & 52.2 & 62.6  \\
    \checkmark &  \checkmark &\checkmark  &\checkmark & \checkmark & 53.0 & 63.8 \\
    \bottomrule
    \end{tabular}
}
\label{tab:loss_ablations_v3}
\end{table}

\begin{table}[t]
    \centering
    \caption{
        Experimental results on ScanNetV2 and Matterport3D in terms of \textit{val} on linear probing evaluation. Upperbound-full sup. denotes the fully-supervised upperbounding results while Baseline init. means initialize the model from our baseline model and then perform linear probing evaluation. Adding AMRADIO for comparision.
        }
        \vspace{-0.1cm}
    \tabcolsep 4pt
    \resizebox{.99\columnwidth}{!}{%
    \begin{tabular}{l|l|cc}
        \toprule
        Type & \textbf{Method} & \multicolumn{2}{c}{\textbf{ScanNetV2}} \\
        ~ & ~ & mIoU & mAcc \\
        \midrule
        Upperbound-fully sup. & MinkowskiNet~\cite{choy20194d} & 68.9 & 77.4 \\
        Baseline init. & MinkowskiNet~\cite{choy20194d} & 54.4 & 64.7 \\
        \midrule

        Concat (Lseg+DINOv2+StableDiffusion) & 3-heads concat & 62.1 & 72.7 \\
        \midrule
        Concat (Lseg+DINOv2+StableDiffusion+AMRADIO) & 4-heads concat & 62.1 & 72.9 \\
        \bottomrule
    \end{tabular}
    }
    \label{tab:linear_probe_supple}
\end{table}

\begin{table}[!tb]
\label{ablationLoss}
\caption{
    Ablation: study on different open-vocabulary 2D semantic segmentation model supervision.
}
\vspace{-0.2cm}
\centering
\resizebox{.99\columnwidth}{!}{%
    \begin{tabular}{ccc|cc}
    \toprule
    \makecell*[c]{Lseg} & OpenSeg & Lseg+OpenSeg  & mIoU $\uparrow$ & mAcc $\uparrow$\\
     \midrule
    \checkmark & ~ & ~ & \textbf{51.4} & 62.3 \\
    ~  & \checkmark & ~ & 44.0 & 62.8 \\
    ~ & ~ &\checkmark & 48.5 & 62.8  \\
    \bottomrule
    \end{tabular}
}
\label{tab:lseg_openseg}
\end{table}

\section{Comparisons with AMRADIO}
\label{sec:amradio_com}

 As the very recent work AMRADIO~\cite{ranzinger2024radio} proposes to distill knowledge from various 2D foundation models (including CLIP, DINOv2, and SAM) which act as teachers, we also employ extra experiments to ablate whether combining supervisions from AMRADIO model can help boost the 3D model distillation. As shown in Table~\ref{tab:loss_ablations_v3}, after extracting the 2D multi-view posed image embeddings and projecting them into the corresponding 3D space, we train the 3D model with feature supervisions from Lseg, DINOv2, Stable Diffusion and AMRADIO simultaneously. However, we discover this training will bring a slight performance drop, from 52.7\% mIoU in our main paper to 52.2\% mIoU in terms of the open-vocabulary semantic segmentation evaluation on ScanNetV2 \textit{val} set. Furthermore, we also extend the proposed deterministic uncertainty estimation for this study, but no further improvement can be obtained. We analyze this is because AMRADIO has been constructed as a student model to acquire knowledge from 2D foundation teacher models, like CLIP, DINOv2, and SAM, resulting in there are no extra informative language or geometric knowledge from the student model, AMRADIO, even causing noisy supervision for the 3D model distillation. Additionally, a linear probing experiment is conducted in Table~\ref{tab:linear_probe_supple}, there are no further improvements from the 3D model which is distilled from four 2D foundation models in parallel.

\section{Lseg and OpenSeg Supervision Combination}
\label{sec:lseg_openseg}

 In this section, we provide additional studies about combing both Lseg and OpenSeg 2D models, being trained to align with the text embeddings for the 2D image open-vocabulary semantic segmentation task. As shown in Table~\ref{tab:lseg_openseg}, a significant performance drop can be observed in terms of open-vocabulary semantic segmentation validation on ScanNetV2 \textit{val} set, though both Lseg and OpenSeg have been aligned with the language space on dense-level supervisions. Since Lseg and OpenSeg present different mask segmentation results for pixels, especially encountering complicate contexts from rapidly changing posed images, this leads to confusion for the 3D model training, attempting to ``cluster'' the same local regions into different `clusters'. Hence, we prefer to utilizing Lseg in our main paper, which is also based on the CLIP.

\section{Parameter $\sigma$ Evolution}
\label{sec:sigma_vis}

\begin{figure*}[tb]
  \centering
  \includegraphics[width=0.8\textwidth]{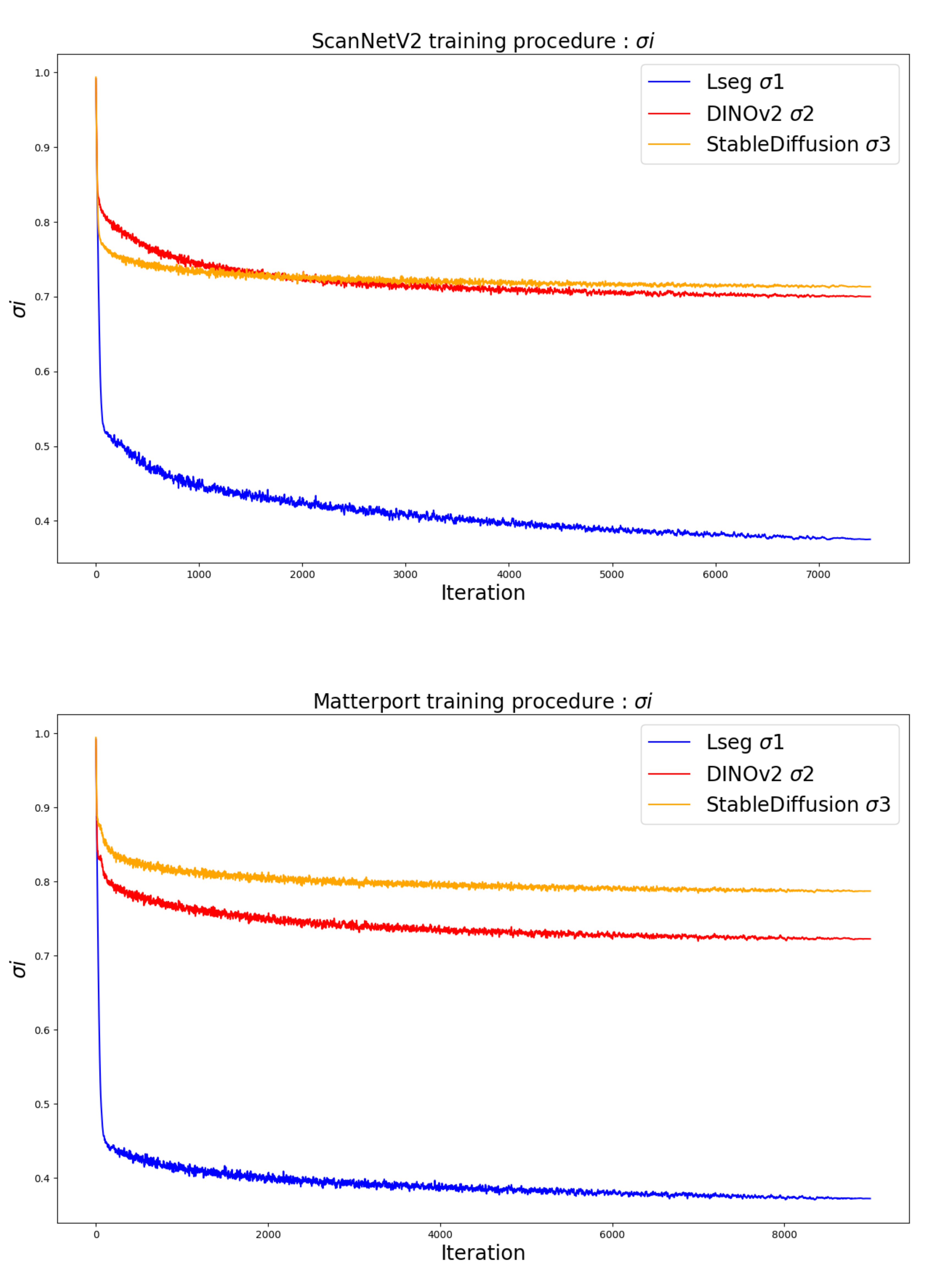}
  \caption{
      Evolutions of parameters in terms of deterministic uncertainty estimation $\sigma_i$.
  }
  \label{fig:aml_params}
\end{figure*}

As we validate in Table 6 in our main paper, we find that employing Auto-Weighting learning results in a trivial solution, we then provide the evolutions of parameters in terms of deterministic uncertainty estimation $\sigma_i$ to help with better capturing the learning using our method CUA-O3D during the distillation in Fig.~\ref{fig:aml_params}.

\section{More Visualizations}
\label{sec:moremore_vis}

\begin{figure*}[tb]
  \centering
  \includegraphics[width=0.99\textwidth]{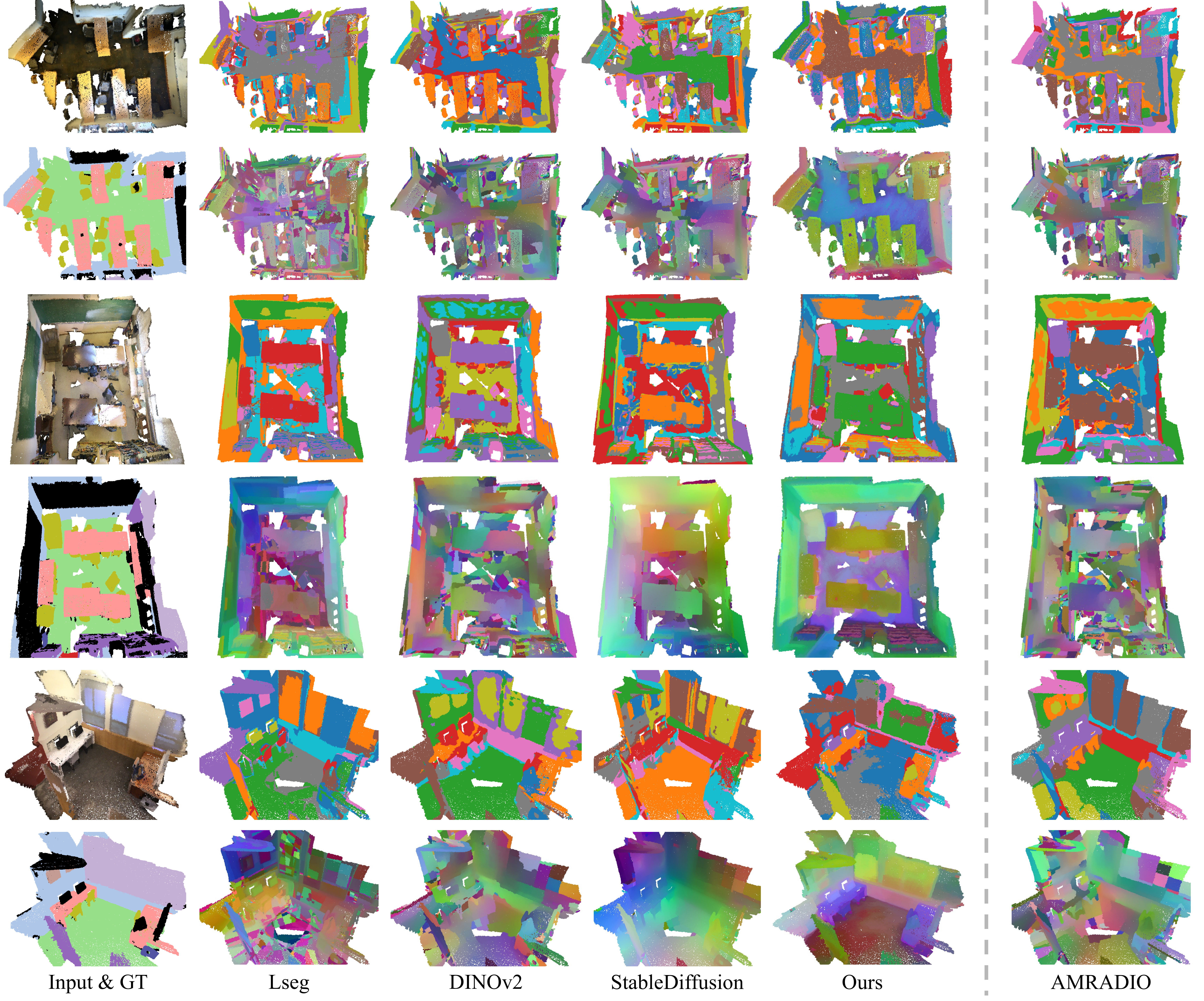}
  \caption{
      Clustering visualizations of ScanNetV2 from various 2D foundation models.
  }
  \label{fig:km_scannet}
\end{figure*}

\begin{figure*}[tb]
  \centering
  \includegraphics[width=0.99\textwidth]{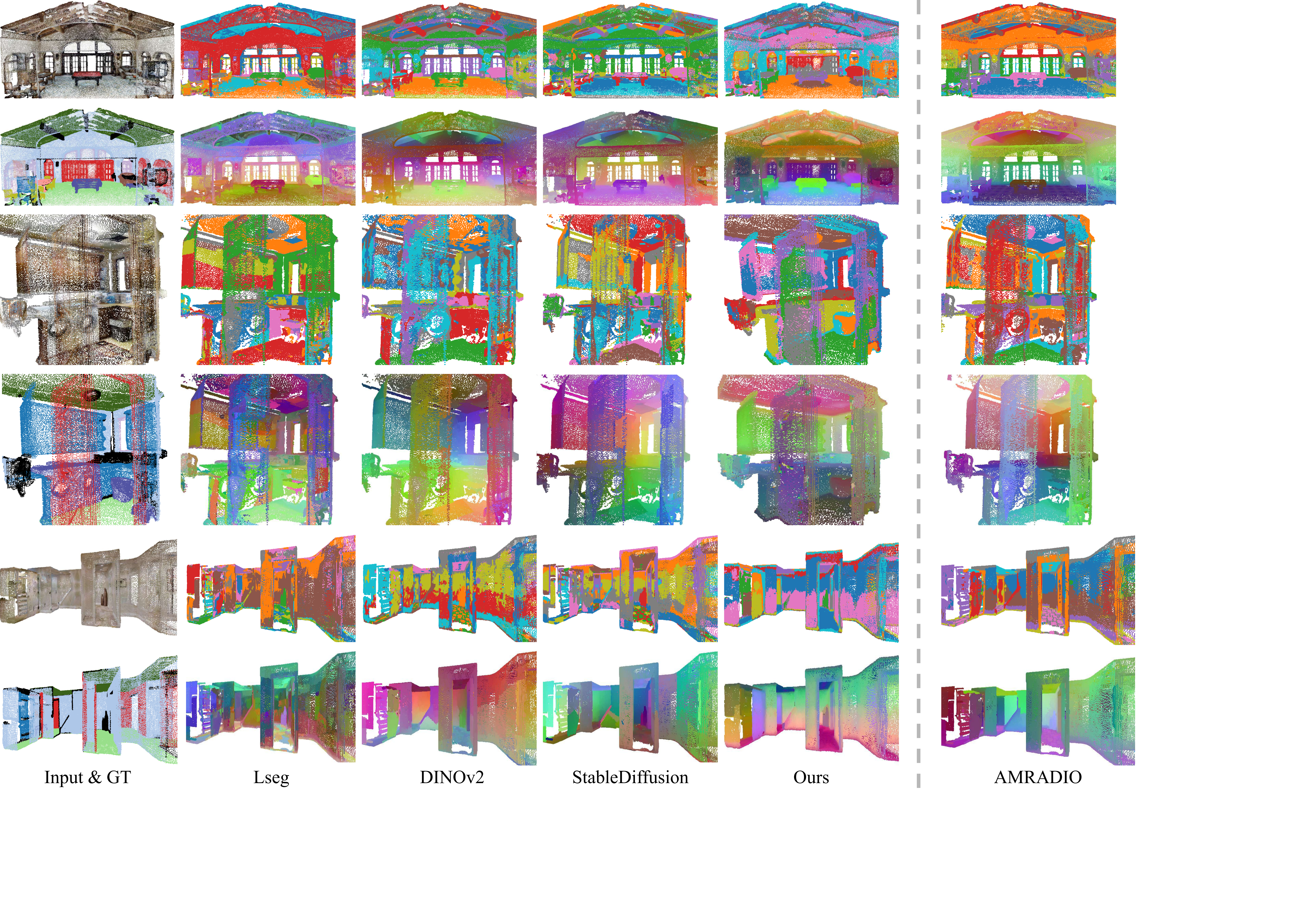}
  \caption{
      Clustering visualizations of Matterport3D from various 2D foundation models.
  }
  \label{fig:km_matterport}
\end{figure*}

\begin{figure*}[tb]
  \centering
  \includegraphics[width=0.99\textwidth]{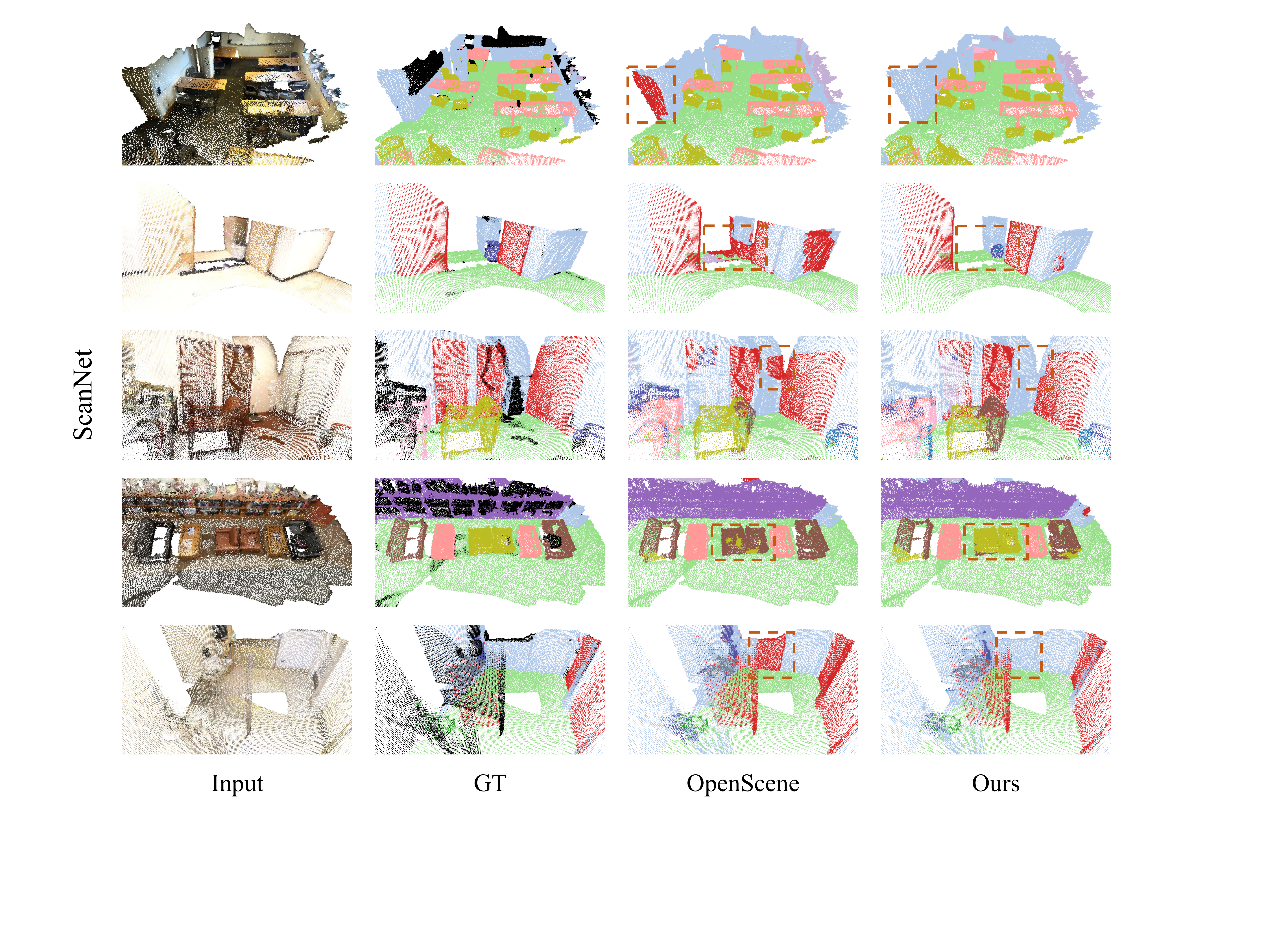}
  \caption{
      Open-vocabulary semantic segmentation visualizations of ScanNetV2.
  }
  \label{fig:scannet_seg}
\end{figure*}

\begin{figure*}[tb]
  \centering
  \includegraphics[width=0.99\textwidth]{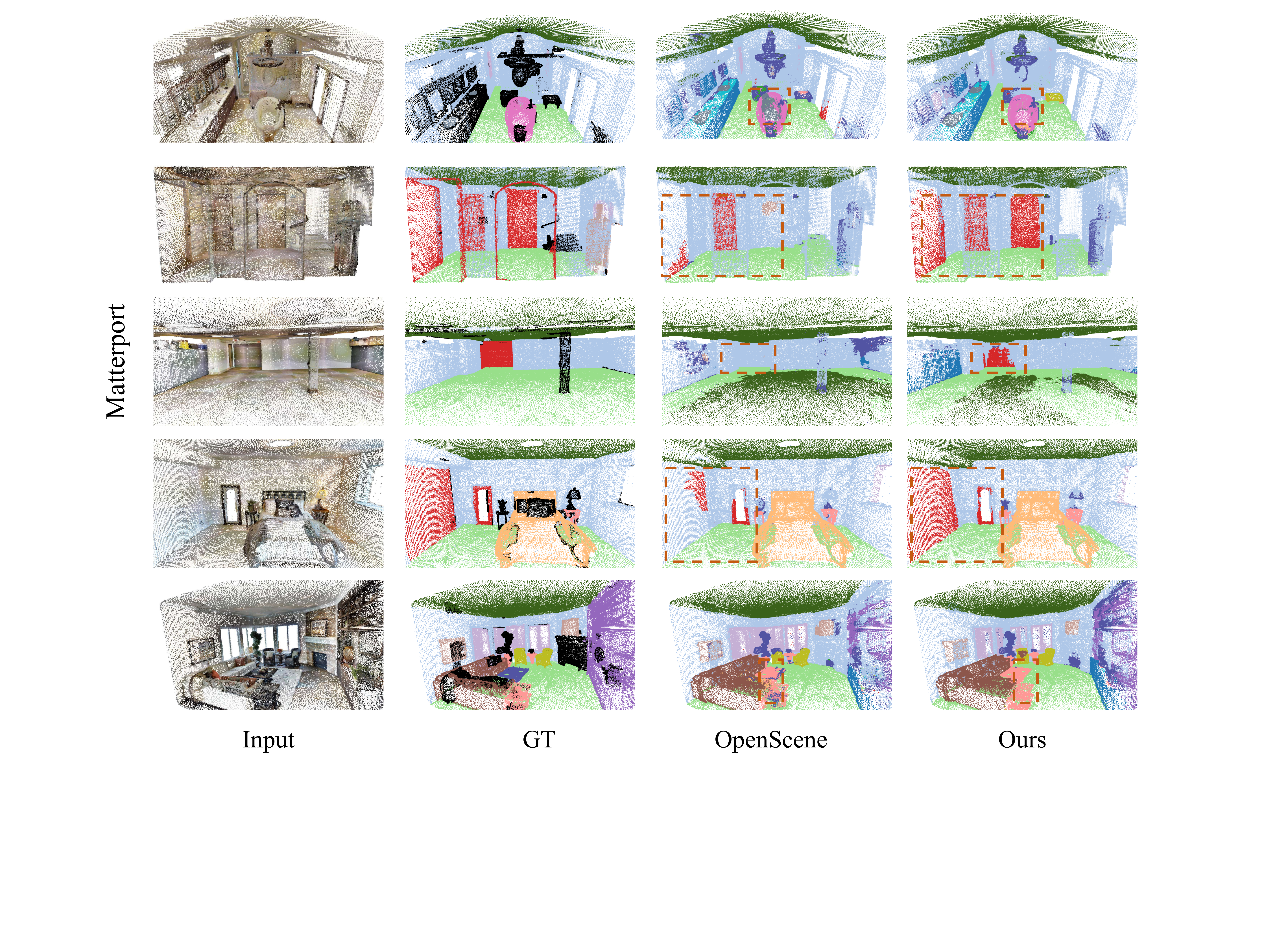}
  \caption{
      Open-vocabulary semantic segmentation visualizations of Matterport3D.
  }
  \label{fig:matterport_seg}
\end{figure*}

In this section, we demonstrate more clustering results for ScanNetV2 and Matterport3D to further clarify the motivation of agglomerating heterogeneous and complementary feature supervisions from various 2D foundation models in Fig.~\ref{fig:km_scannet} and Fig.~\ref{fig:km_matterport}, while open-vocabulary semantic segmentation visualizations are displayed in Fig.~\ref{fig:scannet_seg} and Fig.~\ref{fig:matterport_seg}.

\section{Limitations and Future Improvements} 
\label{sec:limitation}

Our method CUA-O3D inspires three aspects about the potential limitations and future improvements, which can serve as starting points for future research. Firstly, the 3D model distilled from 2D Vision-Language Models shows significantly lower performance than the fully supervised baselines, necessitating more explorations to boost the performance and bridge this gap. Secondly, though our distilled model presents essential improvements over the baseline, there is still no in-depth study about the alignment between visual embedding (3D) and text embedding. Thirdly, how to naturally transform the backbone architectures from 2D to 3D within the multi-view 3D scene understanding shall be interesting, which inherits the strong generalizability and zero-shot capacities from 2D foundation models, while also exploring weakly/semi-supervised setting~\cite{li2022expansion,li2022weakly,zhang2024video,zang2024generalized} under the in-context learning or inducing video sequences data~\cite{zhou2024mlvu,shu2024video} to help the model better understand the 3D scene to further enhance the 3D model leaves interesting insights.

\clearpage
\newpage

\label{sec:more_vis}



\end{document}